\documentclass{article}
\usepackage{spconf,amsmath,graphicx,hyperref,multirow,pifont, subcaption}

\title{WaterFlow: Explicit Physics-Prior Rectified Flow for Underwater Saliency Mask Generation}
\name{Runting Li$^{\rm 1*}$\thanks{*Equal Contribution.} \qquad Shijie Lian$^{\rm 2*}$\footnotemark[1] \qquad Hua Li$^{\rm 1 \dag}$\thanks{$\dag$ Corresponding author.} \qquad Yutong Li$^{\rm 1}$ \qquad Wenhui Wu$^{\rm 3}$ \qquad Sam Kwong$^{\rm 4}$}
  
\address{$^{1}$Hainan University, China  \quad
    $^{2}$Huazhong University of Science and Technology, China\\
    $^{3}$Shenzhen University, China \quad
    $^{4}$Lingnan University, Hong Kong }

\begin{document}
\ninept
\maketitle

\begin{abstract}
Underwater Salient Object Detection (USOD) faces significant challenges, including underwater image quality degradation and domain gaps. Existing methods tend to ignore the physical principles of underwater imaging or simply treat degradation phenomena in underwater images as interference factors that must be eliminated, failing to fully exploit the valuable information they contain. We propose WaterFlow, a rectified flow-based framework for underwater salient object detection that innovatively incorporates underwater physical imaging information as explicit priors directly into the network training process and introduces temporal dimension modeling, significantly enhancing the model's capability for salient object identification. On the USOD10K dataset, WaterFlow achieves a \textbf{0.072} gain in \(S_m\), demonstrating the effectiveness and superiority of our method. \url{https://github.com/Theo-polis/WaterFlow}.
\end{abstract}

\section{Introduction}
\label{sec:intro}
Underwater salient object detection aims to automatically identify and localize the most important or attention-grabbing objects in underwater images or videos \cite{usod10k}. Due to its significant applications in marine biology research, underwater robot navigation, environmental monitoring, and ocean exploration, it has been attracting increasing attention \cite{Lian}\cite{svam}\cite{lian2024diving}.

Currently, thanks to large-scale datasets and high-quality annotations, the field of salient object detection in terrestrial scenarios has made remarkable progress \cite{land1}\cite{land2}\cite{land3}. However, its counterpart in underwater environments faces considerable challenges, mainly due to the following reasons: 1) underwater images captured by devices often suffer from various types of degradation, such as color distortion, detail blurring, and contrast loss, which weaken the saliency of objects and hinder accurate foreground localization; 2) the marine environment differs greatly from terrestrial environments, exhibiting obvious domain shifts, and thus the direct transfer of terrestrial salient object detection techniques to underwater scenes often leads to unsatisfactory performance \cite{Lian}\cite{seathru}.

In recent years, generative models, particularly diffusion models \cite{ddpm}, have demonstrated remarkable performance across a variety of computer vision tasks \cite{waterdiffusion}\cite{camodiffusion_conf}. Diffusion models, by learning the generative process of data distributions, are capable of handling uncertainties within the data and producing high-quality and diverse samples, which hold significant potential for underwater salient object detection. However, applying diffusion models directly to underwater object detection is hindered by computational inefficiency and unstable training. Traditional diffusion models typically require a large number of sampling steps to produce high-quality results, which is impractical for real-time underwater applications.

Moreover, many deep learning-based segmentation methods mainly focus on network architectures and feature modeling from a general computer vision perspective \cite{u2net}\cite{minet}\cite{bbrf}. The imaging process follows complex physical principles, including dual attenuation from direct transmission and backscattering, wavelength-dependent scattering, and depth-induced color distortion \cite{seathru}. These processes not only degrade image quality but also contain rich information about scene structure and object characteristics.
\begin{figure}
    \centering
    \includegraphics[width=1\linewidth]{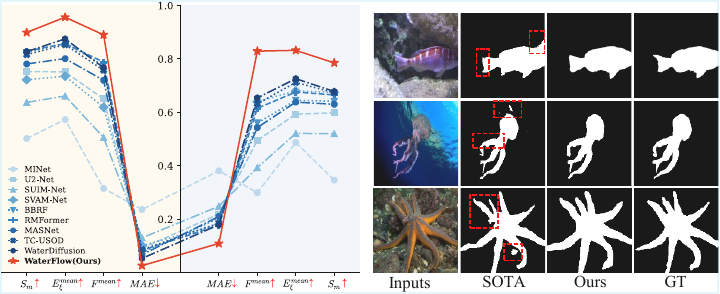}
    \caption{The left panel shows the performance of our model on the \textbf{USOD10K}(L) and \textbf{UFO-120}(R) datasets, while the right panel compares our method against the previous SOTA approaches \cite{waterdiffusion}. Red boxes highlight segmentation errors.}
    \label{fig:1}
\end{figure}
\begin{figure*}[!h]
    \centering
    \includegraphics[width=1\linewidth]{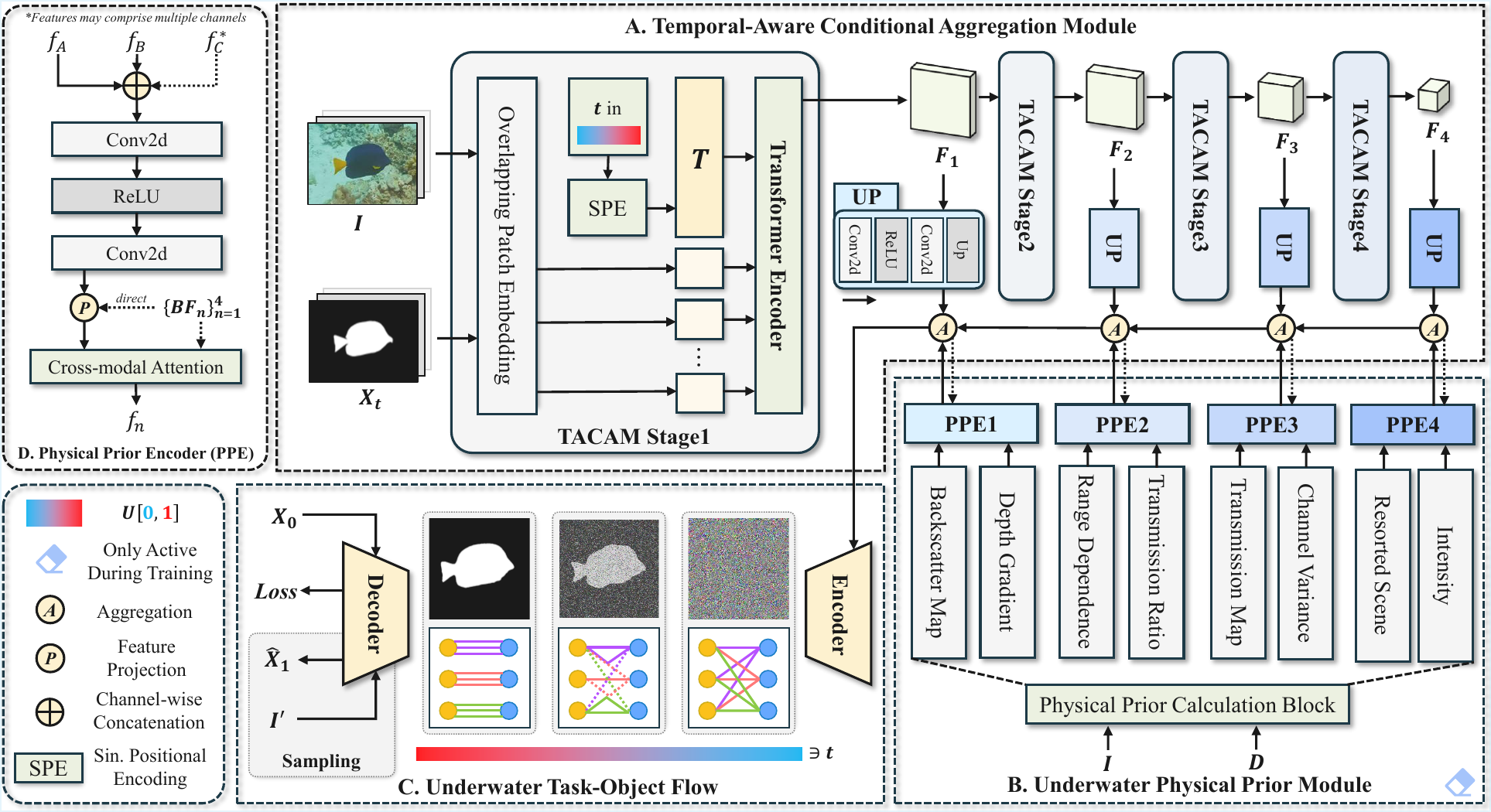}
    \caption{\textbf{WaterFlow Architecture}. The Temporal-Aware Conditional Aggregation Module (TACAM, defined in Sec. \ref{sec:TACAM}) and the Underwater Physical Prior Module (UPPM, defined in Sec. \ref{sec:UPPM}) jointly extract multi-scale features from the input images and masks, which are then fused to guide the downstream Underwater Task Object Flow (UTOF, defined in Sec. \ref{sec:UTOF}) in generating saliency segmentation masks.}
    \label{fig:2}
\end{figure*}
Some studies have attempted to leverage these physical principles by treating image degradations as noise to be removed \cite{semantic}\cite{svam}. Although this can yield visually clearer images, it may also discard valuable information and introduce artificial artifacts. Other approaches adopt multi-task learning frameworks to jointly optimize enhancement and segmentation \cite{waterdiffusion}, but the discrepancy between the two objectives often leads to optimization challenges and increased model complexity. Inspired by these observations, we argue that the underwater imaging process itself encodes valuable priors which, if properly utilized, can enhance both learning efficiency and generalization. To this end, we construct novel physical prior feature maps and explicitly introduce them as an independent modality into the feature space, while maintaining a single-task objective for saliency detection.

Based on the aforementioned challenges and analysis, we propose WaterFlow, a rectified flow-based framework for underwater salient mask generation. WaterFlow can more comprehensively extract spatial semantics from images and endow them with temporal characteristics. By explicitly fusing underwater physical imaging priors with image features, it constructs more comprehensive feature representations, thereby enhancing the performance of SOD tasks. We adopt the Rectified Flow \cite{rectifiedflow} framework to leverage its advantages of linear trajectory learning, achieving more efficient sampling and more stable training processes.
Our main contributions are summarized as follows:
\begin{itemize}
\item We design a conditional-guided underwater saliency detection model named WaterFlow, which innovatively incorporates underwater physical imaging information as explicit prior knowledge directly into the network training process.

\item We are the first to systematically apply Rectified Flow to the challenging field of underwater salient object detection and introduce effective temporal dimension modeling strategies to efficiently meet the diverse feature requirements of different generative stages.

\item WaterFlow achieves substantial performance improvements on USOD10K and UFO-120 datasets, reaching state-of-the-art results, which validate its effectiveness and superiority.
\end{itemize}

\section{Related Work}
\label{sec:format}
\noindent\textbf{CNN and Transformer-based Segmentation Paradigms.} Early methods relied on convolutional neural networks, achieving progress in terrestrial scenarios through multi-scale feature extraction and boundary refinement \cite{u2net}\cite{minet}\cite{bbrf}\cite{masnet}\cite{suimnet}. With the development of self-attention, Transformers were introduced into saliency segmentation and gradually became a new paradigm \cite{rmformer}\cite{usod10k}. However, these methods generally ignore underwater imaging physics, and their performance is limited under severe degradations and domain shifts.

\noindent\textbf{Diffusion-based Generative Paradigms.} In recent years, diffusion models have shown outstanding performance in image generation, restoration, and segmentation, becoming a major direction in generative modeling \cite{waterdiffusion}. Their step-by-step denoising process produces high-quality results, but low sampling efficiency and unstable training make them unsuitable for real-time and robust underwater detection. Moreover, the incorporation of physical priors remains an open problem.

\section{Method}
\label{sec:pagestyle}
\subsection{Overall Architecture}
We regard SOD as a conditional generation task of saliency masks \cite{camodiffusion_conf}, and accordingly design WaterFlow. As illustrated in Fig.~\ref{fig:2}, the Temporal-Aware Conditional Aggregation Module (TACAM) progressively extracts multi-scale RGB features through a four-level pyramid structure and fuses them with coarse masks. Meanwhile, the Underwater Physical Prior Module (UPPM), guided by depth information, captures underwater physical priors such as light attenuation, color shift, and scattering, and progressively integrates them with backbone features through a hierarchical strategy. Finally, the downstream Underwater Task Object Flow (UTOF) takes the multimodal feature maps as conditional input and generates high-quality saliency masks.
\subsection{Temporal-Aware Conditional Aggregation Module}\label{sec:TACAM}
TACAM (Fig.~\ref{fig:2}A) adopts an improved Pyramid Vision Transformer \cite{pvtv2} as the backbone network to extract sufficiently discriminative features, addressing the complex relationship between underwater salient objects and background regions, and obtains the initial spatial embedding \(\{OP_{n}\}_{n=1}^{4}\). Specifically,

\begin{equation}
   OP_n = 
   \begin{cases} 
       Conv2d(R(Conv2d(I) + Conv2d(X_t))), & n = 1 \\
       Conv2d(R(Conv2d(OP_{n-1}))), & n = 2, 3, 4 
   \end{cases}
   \label{eq:op_n}
\end{equation}
where $R(\cdot)$ denotes the reshape operation.

Traditional diffusion-based methods usually employ fixed features during the reverse process \cite{TA}. This static strategy leads to insufficient temporal adaptability of the model: the early steps rely more on global semantic understanding to establish coarse underwater object contours, while the later steps require fine local details to refine boundaries. Inspired by previous work \cite{camodiffusion_conf}, we introduce temporal information as a separate feature embedded into the network:
\begin{equation}
    T_{n}=SPE(t), t\sim U[0,1]
    \label{eq:1}
\end{equation}
Subsequently, the temporal embedding is concatenated with the spatial embedding and processed through Transformer blocks to obtain \(\{F_{n}\}_{n=1}^{4}\). \(F_{n}\) passes through UP blocks to finally obtain backbone features \(\{BF_{n}\}_{n=1}^{4}\). However, underwater scenes present greater complexity than terrestrial environments. To address these domain-specific characteristics, we introduce a dedicated Underwater Physical Prior Module that enhances feature representation by incorporating water-specific knowledge.


\subsection{Underwater Physical Prior Module}\label{sec:UPPM}
Light propagation in underwater environments follows complex physical laws \cite{seathru}. Directly performing saliency segmentation on degraded underwater images often leads to performance degradation when facing complex underwater scenes.

Depth information, as a core element in underwater physical modeling, plays a crucial role in characterizing light propagation and attenuation. Therefore, for the training set \(I\), we employed the state-of-the-art monocular depth estimation model, \textit{Depth Anything V2} \cite{depth_anything_v2}, to obtain the corresponding depth maps \(D\). The test set remains unprocessed, as this module is only activated during training.

Existing approaches that leverage physical priors mainly follow two strategies: one is to perform complex enhancement preprocessing on raw underwater images and then apply salient object detection on the enhanced images \cite{semantic}\cite{svam}; the other is to treat underwater salient object detection and image restoration as parallel tasks and formulate them as a cooperative optimization process \cite{waterdiffusion}. In our view, the former design suffers from information loss and error accumulation, as the separation between enhancement and saliency detection may distort or discard critical details relevant to saliency. Although the latter avoids the drawbacks of serial processing, the inherent inconsistency between the two task objectives often leads to optimization conflicts, preventing physical priors from fully serving the core requirements of saliency detection, while also introducing unnecessary network complexity.

In contrast, we innovatively incorporate physical information as explicit prior knowledge directly into the network, thereby avoiding intermediate information loss and enabling physical constraints to guide saliency determination in a more elegant and precise manner.(Fig.~\ref{fig:2}B)

Specifically, we model the relationship between underwater captured images and their true versions as: 
\begin{equation}
    I_c(x) = J_c(x) \cdot T_c^D(x) + A_c \cdot (1 - T_c^B(x))
    \label{eq:2}
\end{equation}
where \(I_c(x)\) represents the pixel intensity of the input image at channel c, and \(J_c(x)\) represents its true version. \(A_c\) is the background light, whose estimation is based on the dark pixel assumption. \(T_c^D(x)\) and \(T_c^B(x)\) represent direct transmission and backscattering transmission, respectively \cite{seathru}.

We also extract several common physical features and construct a physical feature set: 
\begin{equation}
    f=\left\{B_c\left(x\right),\ \nabla z\left(x\right),\beta_c^D,R,T_c^D\left(x\right),{Var}_c,J_c\left(x\right),{Int}\right\}
    \label{eq:3}
\end{equation}
where \(B_c(x)\) represents the backscattering map, \(\nabla z(x)\) represents depth gradient, \(\beta_c^D\) represents Range-Dependent Attenuation Coefficient, \(R\) represents cross-channel transmission ratio, \(Var_c\) represents channel variance, and \(Int\) represents backscattering intensity and attenuation intensity maps.

Physical priors injected only at a single scale are difficult to impose effective constraints on both low-level details and high-level semantics simultaneously. Therefore we hierarchically encode physical priors according to semantics: 
\begin{equation}
    f_n = {PPE}_n\left(f_{i|n}\right), \quad n=1, 2, 3, 4; \quad i=A, B, C
    \label{eq:4}
\end{equation}
where \(f_{i|n}\in f\) represents features \(i\) participating in encoding stage \(n\). \(PPE_n(\cdot)\) represents the n-th Physical Prior Encoder, whose specific architecture is described in Fig.~\ref{fig:2}D. Shallow-layer encoding introduces boundary-related physical information such as backscattering and depth gradients, middle layers fuse distance-varying attenuation characteristics and channel difference information, while deep layers impose globally-related physical constraints.
\begin{table*}[!t]
\centering
\begin{tabular}{c|c|cccc|cccc}
\hline
\hline
\multirow{2}{*}{Method} & \multirow{2}{*}{Pub.} & \multicolumn{4}{c|}{USOD10K} & \multicolumn{4}{c}{UFO-120} \\
\cline{3-10}
& & \(MAE\) $\downarrow$ & \(F^{mean}\) $\uparrow$ & \(E_\xi^{mean}\) $\uparrow$ & \(S_m\) $\uparrow$ & \(MAE\) $\downarrow$ & \(F^{mean}\) $\uparrow$ & \(E_\xi^{mean}\) $\uparrow$ & \(S_m\) $\uparrow$ \\
\hline
\hline
MINet \cite{minet} & CVPR-20 & 0.236 & 0.315 & 0.573 & 0.502 & 0.381 & 0.299 & 0.487 & 0.346 \\
U2-Net \cite{u2net} & PR-20 & 0.091 & 0.650 & 0.751 & 0.753 & 0.219 & 0.495 & 0.593 & 0.598 \\
SUIM-Net \cite{suimnet} & IROS-20 & 0.130 & 0.506 & 0.661 & 0.637 & 0.248 & 0.393 & 0.521 & 0.520 \\
SVAM-Net \cite{svam} & RSS-22 & 0.101 & 0.619 & 0.735 & 0.722 & 0.180 & 0.640 & 0.680 & 0.675 \\
BBRF \cite{bbrf} & TIP-23 & 0.074 & 0.776 & 0.859 & 0.823 & 0.203 & 0.560 & 0.645 & 0.641 \\
RMFormer \cite{rmformer} & MM-23 & 0.083 & \underline{0.788} & 0.857 & \underline{0.829} & \underline{0.177} & 0.615 & 0.676 & 0.665 \\
TC-USOD \cite{usod10k} & TIP-23 & 0.086 & 0.758 & 0.853 & 0.816 & 0.185 & 0.632 & 0.712 & 0.673 \\
MASNet \cite{masnet} & JOE-24 & 0.067 & 0.720 & 0.801 & 0.781 & 0.210 & 0.543 & 0.638 & 0.630 \\
WaterDiffusion \cite{waterdiffusion} & AAAI-25 & \underline{0.053} & 0.767 & \underline{0.875} & 0.827 & 0.179 & \underline{0.654} & \underline{0.726} & \underline{0.678} \\
\textbf{WaterFlow(Ours)} &  & \textbf{0.026} & \textbf{0.890} & \textbf{0.956} & \textbf{0.899} & \textbf{0.109} & \textbf{0.829} & \textbf{0.832} & \textbf{0.785} \\
\hline
\hline
\end{tabular}
\caption{Quantitative evaluation of various methods across different public underwater datasets, where WaterDiffusion serves as the previous state-of-the-art. The \textbf{best} and \underline{second-best} results are highlighted with bold and underlined, respectively.}
\label{tab:1}
\end{table*}
\begin{figure*}
    \centering
    \includegraphics[width=1\linewidth]{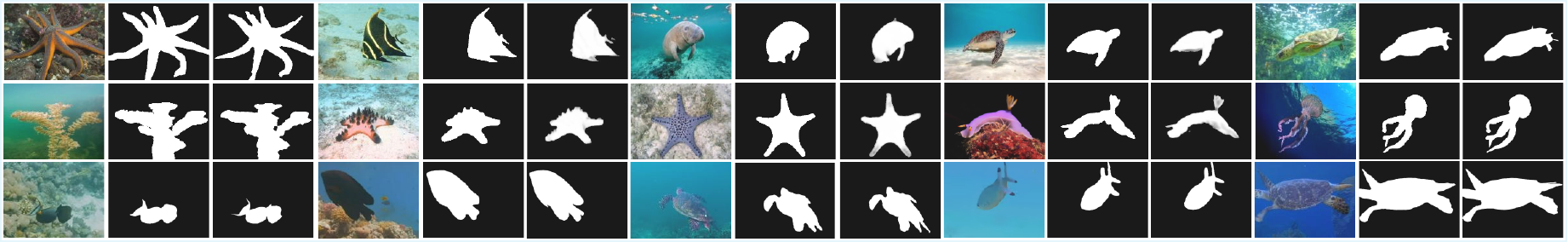}
    \caption{Qualitative results of saliency detection for our method. In each group of images, the one on the left is the input image, the middle is the segmentation result generated by our model, and the one on the right is the ground truth (GT).}
    \label{fig:3}
\end{figure*}
\begin{table}[htbp]
\centering
\begin{subtable}{0.48\linewidth}
  \centering
  \begin{tabular}{cc|cc}
\hline
\hline
\multicolumn{2}{c|}{Module} & \multirow{2}{*}{\(MAE\)}  & \multirow{2}{*}{\(S_m\)}\\
\cline{1-2}
TA & UPPM & \\
\hline
\hline
\ding{55} & \ding{51} & 0.030 & 0.884 \\
\ding{51} & \ding{55} & 0.034 & 0.874 \\
\hline
\hline 
\end{tabular}
  \caption{Component Ablation.}
  \label{tab:2.a}
\end{subtable}
\hspace{0.02\linewidth} 
\begin{subtable}{0.48\linewidth}
  \centering
  \begin{tabular}{c|cc}
\hline
\hline
    Method & Steps & FPS \\
\hline
    SOTA \cite{waterdiffusion} & 1000 & 2.4 \\
    \textbf{Ours} & \textbf{1} & \textbf{20} \\
\hline
\hline
\end{tabular}
  \caption{Time Performance.}
  \label{tab:2.b}
\end{subtable}
\caption{Ablation Study on the USOD10K Dataset.}
\label{tab:2}
\end{table}

\subsection{Underwater Task Object Flow}\label{sec:UTOF}
In the downstream stage, WaterFlow employs an improved conditional generative model based on continuous normalizing flows \cite{rectifiedflow}. Unlike conventional diffusion-based baselines \cite{ddpm}, we aim to learn a parameterized conditional vector field network \(v_\theta(X_t,t,I)\), which takes as input the position \(X_t\), the time step \(t\), and the condition information \(I\). By solving the ODE \(\frac{dX_t}{dt}=v_\theta(X_t,t,I)\), the initial noise distribution \(\pi_0\) is transformed into the target data distribution \(\pi_1(X_t|I)\), i.e., the conditional distribution of the real data \(X_t\) given condition \(I\).(Fig.~\ref{fig:2}C)

Formally, given \(X_0\sim\pi_0\) sampled from the noise distribution and \(X_1\sim\pi_1(X_t|I)\) sampled from the conditional data distribution, we define a straight-line interpolation path: 
\begin{equation}
    X_t = tX_1 + (1-t)X_0
    \label{eq:6}
\end{equation}
where \(\forall t\in\left[0,1\right]\). However, Simply using L2 loss pays insufficient attention to boundary accuracy and structural integrity in segmentation tasks \cite{camodiffusion_conf}. Therefore, we adopt Task Loss as the optimization objective for the network: 
\begin{equation}
Loss\left(\theta\right)=\frac{1}{2}L_{BCE}\left(\theta\right)+\frac{1}{2}L_{IoU}\left(\theta\right)
\label{eq:7}
\end{equation}
Task Loss combines binary cross-entropy and IoU loss, addressing the challenge of underwater object boundary learning by assigning higher weights to boundary regions. 

\section{Experiments}
\label{sec:typestyle}
\subsection{Experiment Settings}
\textbf{Datasets.} We use the USOD10K dataset \cite{usod10k} as the training set to train our model and conduct a comprehensive evaluation on the USOD10K and UFO-120 \cite{ufo120} dataset.

\noindent\textbf{Evaluation metrics.} To comprehensively evaluate the proposed model, we adopt popular metrics for evaluation, including F-measure \cite{Fm}, S-measure \cite{Sm}, E-measure \cite{Em}, and MAE \cite{MAE}.

\noindent\textbf{Implementation details.} We implement our model based on the PyTorch framework, using a single NVIDIA RTX 4090 GPU for training and inference. TACAM is initialized with PVTv2-B4 \cite{pvtv2}, with input images of 352×352 resolution. We use the AdamW optimizer with a batch size of 8 and gradient accumulation steps of 4. The learning rate is set to 2.5e-5 and training is conducted for 90 epochs. The sampling step is set to 1.
\subsection{Evaluation and Ablation Study}
\noindent\textbf{Quantitative Evaluation.} We compared WaterFlow with nine representative methods on public benchmarks, and the results are shown in Tab.~\ref{tab:1}. It can be observed that WaterFlow achieves the best performance across all metrics and significantly exceeds the previous state-of-the-art method \cite{waterdiffusion}.

\noindent\textbf{Qualitative Evaluation.} We provided a comparison with WaterDiffusion in Fig.~\ref{fig:1}, where its results show foreground misactivation or boundary deviations in certain regions, while WaterFlow produces more accurate and coherent masks. Furthermore, Fig.~\ref{fig:3} presents comparisons between WaterFlow and ground truth, demonstrating that our model can consistently generate high-quality segmentation results across diverse underwater environments.

\noindent\textbf{Ablation and Analysis.} Tab.\ref{tab:2.b}~compares WaterFlow and WaterDiffusion \cite{waterdiffusion} in terms of sampling steps and FPS performance. WaterFlow achieves over 8× improvement in processing speed while maintaining high accuracy.

We believe that the Temporal-Aware mechanism overcomes the limitations of traditional methods that use a single control image from start to finish by providing adaptive feature inputs at each step during the time-step iteration process of Rectified Flow, enabling the model to dynamically adjust feature representations according to the flow requirements at different time steps. UPPM explicitly integrates the physical prior knowledge of underwater imaging into feature maps, providing the model with richer and more accurate physical constraint information. Furthermore, the adoption of Rectified Flow's straight-line interpolation strategy significantly improves inference speed. Compared to the complex sampling trajectories of traditional diffusion models, the straight-line path achieves a more efficient generation process \cite{rectifiedflow}. This efficient inference capability is of great significance for the real-time requirements of underwater tasks, enabling it to meet the needs of processing speed-sensitive practical applications such as underwater robot navigation and marine life monitoring.

\section{Conclusion}
\label{sec:copyright}
We presented WaterFlow, a rectified flow-based model for underwater salient object detection. By explicitly integrating underwater physical priors and temporal modeling, our method achieves more comprehensive feature learning. Experiments on USOD10K and UFO-120 show that WaterFlow outperforms previous SOTA with significant gains. As the first to apply Rectified Flow underwater, we hope that it will be useful for real-time marine work.
\section{Acknowledgments}
This work was supported in part by the National Natural Science Foundation of China under Grant 62461018, 62376162; in part by the Hainan Provincial Natural Science Foundation of China under Grant No. 625YXQN594; in part by the Innovation Platform for "New Star of South China Sea" of Hainan Province under Grant No. NHXXRCXM202306; in part by the Research Grants Council of the Hong Kong Special Administrative Region, China under Grant STG5/E-103/24-R.


\bibliographystyle{IEEEbib}
\bibliography{strings,refs}

@InProceedings{seathru,
author = {Akkaynak, Derya and Treibitz, Tali},
title = {Sea-Thru: A Method for Removing Water From Underwater Images},
booktitle = {CVPR},
month = {June},
year = {2019}
}

@inproceedings{camodiffusion_conf,
  title={CamoDiffusion: Camouflaged object detection via conditional diffusion models},
  author={Chen, Zhongxi and Sun, Ke and Lin, Xianming},
  booktitle={Proceedings of the AAAI Conference on Artificial Intelligence},
  volume={38},
  number={2},
  pages={1272--1280},
  year={2024}
}

@article{rectifiedflow,
  title={Flow straight and fast: Learning to generate and transfer data with rectified flow},
  author={Liu, Xingchao and Gong, Chengyue and Liu, Qiang},
  journal={arXiv preprint arXiv:2209.03003
        },
  year={2022}
}

@inproceedings{waterdiffusion,
  title={WaterDiffusion: Learning a Prior-involved Unrolling Diffusion for Joint Underwater Saliency Detection and Visual Restoration},
  author={Chang, Laibin and Wang, Yunke and Deng, Longxiang and Du, Bo and Xu, Chang},
  booktitle={Proceedings of the AAAI Conference on Artificial Intelligence},
  volume={39},
  number={2},
  pages={1998--2006},
  year={2025}
}

@article{ddpm,
  title={Denoising diffusion probabilistic models},
  author={Ho, Jonathan and Jain, Ajay and Abbeel, Pieter},
  journal={Advances in neural information processing systems},
  volume={33},
  pages={6840--6851},
  year={2020}
}

@inproceedings{lian2024diving,
  title     = {Diving into Underwater: Segment Anything Model Guided Underwater Salient Instance Segmentation and A Large-scale Dataset},
  author    = {Lian, Shijie and Zhang, Ziyi and Li, Hua and Li, Wenjie and Yang, Laurence Tianruo and Kwong, Sam and Cong, Runmin},
  booktitle = {ICML},
  pages     = {29545--29559},
  year      = {2024}
}

@article{pvtv2,
  title={Pvt v2: Improved baselines with pyramid vision transformer},
  author={Wang, Wenhai and Xie, Enze and Li, Xiang and Fan, Deng-Ping and Song, Kaitao and Liang, Ding and Lu, Tong and Luo, Ping and Shao, Ling},
  journal={Computational visual media},
  volume={8},
  number={3},
  pages={415--424},
  year={2022},
  publisher={TUP}
}

@article{ufo120,
  title={Simultaneous enhancement and super-resolution of underwater imagery for improved visual perception},
  author={Islam, Md Jahidul and Luo, Peigen and Sattar, Junaed},
  journal={arXiv preprint arXiv:2002.01155
        
        
        
        
        
        },
  year={2020}
}

@article{usod10k,
  title={Usod10k: a new benchmark dataset for underwater salient object detection},
  author={Hong, Lin and Wang, Xin and Zhang, Gan and Zhao, Ming},
  journal={IEEE transactions on image processing},
  volume={34},
  pages={1602--1615},
  year={2023},
  publisher={IEEE}
}

@inproceedings{Sm,
  title={Structure-measure: A new way to evaluate foreground maps},
  author={Fan, Deng-Ping and Cheng, Ming-Ming and Liu, Yun and Li, Tao and Borji, Ali},
  booktitle={ICCV},
  pages={4548--4557},
  year={2017}
}

@article{Em,
  title={Enhanced-alignment measure for binary foreground map evaluation},
  author={Fan, Deng-Ping and Gong, Cheng and Cao, Yang and Ren, Bo and Cheng, Ming-Ming and Borji, Ali},
  journal={arXiv preprint arXiv:1805.10421
        
        
        
        
        
        
        
        
        
        
        
         },
  year={2018}
}

@inproceedings{Fm,
  title={Frequency-tuned salient region detection},
  author={Achanta, Radhakrishna and Hemami, Sheila and Estrada, Francisco and Susstrunk, Sabine},
  booktitle={CVPR},
  pages={1597--1604},
  year={2009},
  organization={IEEE}
}

@inproceedings{MAE,
  title={Saliency filters: Contrast based filtering for salient region detection},
  author={Perazzi, Federico and Kr{\"a}henb{\"u}hl, Philipp and Pritch, Yael and Hornung, Alexander},
  booktitle={CVPR},
  pages={733--740},
  year={2012},
  organization={IEEE}
}

@article{svam,
  title={SVAM: Saliency-guided visual attention modeling by autonomous underwater robots},
  author={Islam, Md Jahidul and Wang, Ruobing and Sattar, Junaed},
  journal={arXiv preprint arXiv:2011.06252
        
        
        
        
        
        
        
        
        
        
        
        
        
        
        
        
        
        
        
        
        
        
        
        
        
        
        
        
        
        
        
        
        
        },
  year={2020}
}

@article{u2net,
  title={U2-Net: Going deeper with nested U-structure for salient object detection},
  author={Qin, Xuebin and Zhang, Zichen and Huang, Chenyang and Dehghan, Masood and Zaiane, Osmar R and Jagersand, Martin},
  journal={Pattern recognition},
  volume={106},
  pages={107404},
  year={2020},
  publisher={Elsevier}
}

@inproceedings{suimnet,
  title={Semantic segmentation of underwater imagery: Dataset and benchmark},
  author={Islam, Md Jahidul and Edge, Chelsey and Xiao, Yuyang and Luo, Peigen and Mehtaz, Muntaqim and Morse, Christopher and Enan, Sadman Sakib and Sattar, Junaed},
  booktitle={IROS},
  pages={1769--1776},
  year={2020},
  organization={IEEE}
}

@inproceedings{minet,
  title={Multi-scale interactive network for salient object detection},
  author={Pang, Youwei and Zhao, Xiaoqi and Zhang, Lihe and Lu, Huchuan},
  booktitle={CVPR},
  pages={9413--9422},
  year={2020}
}

@article{bbrf,
  title={Boosting broader receptive fields for salient object detection},
  author={Ma, Mingcan and Xia, Changqun and Xie, Chenxi and Chen, Xiaowu and Li, Jia},
  journal={IEEE Transactions on Image Processing},
  volume={32},
  pages={1026--1038},
  year={2023},
  publisher={IEEE}
}

@inproceedings{rmformer,
  title={Recurrent multi-scale transformer for high-resolution salient object detection},
  author={Deng, Xinhao and Zhang, Pingping and Liu, Wei and Lu, Huchuan},
  booktitle={ACM MM},
  pages={7413--7423},
  year={2023}
}

@article{masnet,
  title={Masnet: A robust deep marine animal segmentation network},
  author={Fu, Zhenqi and Chen, Ruizhe and Huang, Yue and Cheng, En and Ding, Xinghao and Ma, Kai-Kuang},
  journal={JOE},
  volume={49},
  number={3},
  pages={1104--1115},
  year={2023},
  publisher={IEEE}
}

@article{semantic,
  title={Semantic segmentation method of underwater images based on encoder-decoder architecture},
  author={Wang, Jinkang and He, Xiaohui and Shao, Faming and Lu, Guanlin and Hu, Ruizhe and Jiang, Qunyan},
  journal={Plos one},
  volume={17},
  number={8},
  pages={e0272666},
  year={2022},
  publisher={Public Library of Science San Francisco, CA USA}
}

@inproceedings{land1,
  title={Scene context-aware salient object detection},
  author={Siris, Avishek and Jiao, Jianbo and Tam, Gary KL and Xie, Xianghua and Lau, Rynson WH},
  booktitle={ICCV},
  pages={4156--4166},
  year={2021}
}

@InProceedings{land2,
    author    = {Wang, Yi and Wang, Ruili and Fan, Xin and Wang, Tianzhu and He, Xiangjian},
    title     = {Pixels, Regions, and Objects: Multiple Enhancement for Salient Object Detection},
    booktitle = {CVPR},
    month     = {June},
    year      = {2023},
    pages     = {10031-10040}
}

@InProceedings{land3,
    author    = {Luo, Ziyang and Liu, Nian and Zhao, Wangbo and Yang, Xuguang and Zhang, Dingwen and Fan, Deng-Ping and Khan, Fahad and Han, Junwei},
    title     = {VSCode: General Visual Salient and Camouflaged Object Detection with 2D Prompt Learning},
    booktitle = {CVPR},
    month     = {June},
    year      = {2024},
    pages     = {17169-17180}
}

@inproceedings{Lian, address={Paris, France}, title={WaterMask: Instance Segmentation for Underwater Imagery}, rights={https://doi.org/10.15223/policy-029}, ISBN={979-8-3503-0718-4}, url={https://ieeexplore.ieee.org/document/10376692/}, DOI={10.1109/ICCV51070.2023.00126}, abstractNote={Underwater image instance segmentation is a fundamental and critical step in underwater image analysis and understanding. However, the paucity of general multiclass instance segmentation datasets has impeded the development of instance segmentation studies for underwater images. In this paper, we propose the first underwater image instance segmentation dataset (UIIS), which provides 4628 images for 7 categories with pixel-level annotations. Meanwhile, we also design WaterMask for underwater image instance segmentation for the first time. In WaterMask, we first devise Difference Similarity Graph Attention Module (DSGAT) to recover lost detailed information due to image quality degradation and downsampling to help the network prediction. Then, we propose Multi-level Feature Refinement Module (MFRM) to predict foreground masks and boundary masks separately by features at different scales, and guide the network through Boundary Mask Strategy (BMS) with boundary learning loss to provide finer prediction results. Extensive experimental results demonstrates that WaterMask can achieve significant gains of 2.9, 3.8 mAP over Mask R-CNN when using ResNet-50 and ResNet-101. Code and Dataset are available at https: //github.com/LiamLian0727/WaterMask.}, booktitle={ICCV}, publisher={IEEE}, author={Lian, Shijie and Li, Hua and Cong, Runmin and Li, Suqi and Zhang, Wei and Kwong, Sam}, year={2023}, month=oct, pages={1305–1315}, language={en} }

@inproceedings{depth_anything_v2,
 author = {Yang, Lihe and Kang, Bingyi and Huang, Zilong and Zhao, Zhen and Xu, Xiaogang and Feng, Jiashi and Zhao, Hengshuang},
 booktitle = {Advances in Neural Information Processing Systems},
 editor = {A. Globerson and L. Mackey and D. Belgrave and A. Fan and U. Paquet and J. Tomczak and C. Zhang},
 pages = {21875--21911},
 publisher = {Curran Associates, Inc.},
 title = {Depth Anything V2},
 url = {https://proceedings.neurips.cc/paper_files/paper/2024/file/26cfdcd8fe6fd75cc53e92963a656c58-Paper-Conference.pdf},
 volume = {37},
 year = {2024}
}

@article{TA,
  title={Segdiff: Image segmentation with diffusion probabilistic models},
  author={Amit, Tomer and Shaharbany, Tal and Nachmani, Eliya and Wolf, Lior},
  journal={arXiv preprint arXiv:2112.00390
        
        
        
        
        
        
        },
  year={2021}
}

\end{document}